\documentclass[letterpaper, 10 pt, conference]{ieeeconf}  

\IEEEoverridecommandlockouts                              

\usepackage{epsfig} 
\usepackage{mathptmx} 
\usepackage{times} 
\usepackage{amsmath} 
\usepackage{amssymb}  
\usepackage{graphicx}
\usepackage{caption}
\usepackage{subcaption}
\usepackage{booktabs}
\usepackage{multirow}
\usepackage{utfsym}
\usepackage{cleveref}
\usepackage{pifont}
 
\title{\LARGE \bf
Structured Labeling Enables Faster Vision-Language Models for End-to-End Autonomous Driving
}
\vspace{-30pt}

\author{Hao Jiang\textsuperscript{1}, Chuan Hu\textsuperscript{1, \ding{41}}, Yukang Shi\textsuperscript{2}, Yuan He\textsuperscript{2}, Ke Wang\textsuperscript{2}, Xi Zhang\textsuperscript{1}, Zhipeng Zhang\textsuperscript{1,\ding{41}}
\thanks{\textsuperscript{1}Shanghai Jiao Tong University, \textsuperscript{2}KargoBot, \textsuperscript{\ding{41}}Corresponding author: \{chuan.hu, zhipengzhang\}@sjtu.edu.cn}
}

\usepackage{titlesec}
\usepackage{indentfirst}
\titlespacing*{\section}{0pt}{4pt}{3pt}
\titlespacing*{\subsection}{0pt}{3pt}{2pt}
\titlespacing*{\subsubsection}{0pt}{2pt}{1pt}

\begin{document}

\maketitle

\begin{abstract}
Vision-Language Models (VLMs) offer a promising approach to end-to-end autonomous driving due to their human-like reasoning capabilities. However, troublesome gaps remains between current VLMs and real-world autonomous driving applications. One major limitation is that existing datasets with loosely formatted language descriptions are not machine-friendly and may introduce redundancy. Additionally, high computational cost and massive scale of VLMs hinder the inference speed and real-world deployment. To bridge the gap, this paper introduces a structured and concise benchmark dataset, NuScenes-S, which is derived from the NuScenes dataset and contains machine-friendly structured representations. Moreover, we present FastDrive, a compact VLM baseline with 0.9B parameters. In contrast to existing VLMs with over 7B parameters and unstructured language processing(e.g., LLaVA-1.5), FastDrive understands structured and concise descriptions and generates machine-friendly driving decisions with high efficiency. Extensive experiments show that FastDrive achieves competitive performance on structured dataset, with approximately 20\% accuracy improvement on decision-making tasks, while surpassing massive parameter baseline in inference speed with over 10$\times$ speedup. Additionally, ablation studies further focus on the impact of scene annotations (e.g., weather, time of day) on decision-making tasks, demonstrating their importance on decision-making tasks in autonomous driving. 
\end{abstract}


\section{Introduction}
\label{sec:intro}
The rapid evolution of autonomous driving systems demands robust environmental understanding capabilities that transcend conventional perception modules \cite{esteban,zhou2024}. The integration of human-like reasoning into autonomous driving systems has become a pivotal research frontier, where Vision-Language Models (VLMs) have emerged as a transformative paradigm, offering human-like reasoning through multimodal fusion of visual inputs and linguistic context. While recent studies have shown the potential of VLMs in scene understanding and decision explanation \cite{xie2025, drivelm, drivevlm, leapad}, critical gaps persist in real-world deployment: inefficient linguistic processing and computational overhead from model scale, which hinder performance and integration into autonomous driving systems.

\begin{figure}
  \setlength{\abovecaptionskip}{0.1cm}
  \centering
  \begin{subfigure}{0.8\linewidth}
    \setlength{\abovecaptionskip}{0.1cm}
      \centering
      \includegraphics[width=\linewidth]{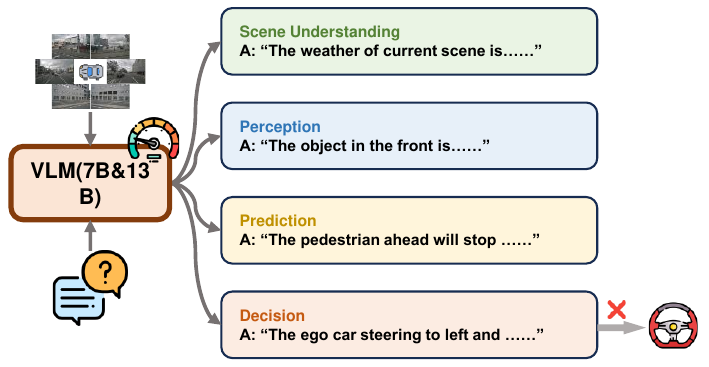}
      \caption{Existing VLMs baseline}
  \end{subfigure}
  \hfill
  \begin{subfigure}{0.8\linewidth}
    \setlength{\abovecaptionskip}{0.1cm}
      \centering
      \includegraphics[width=\linewidth]{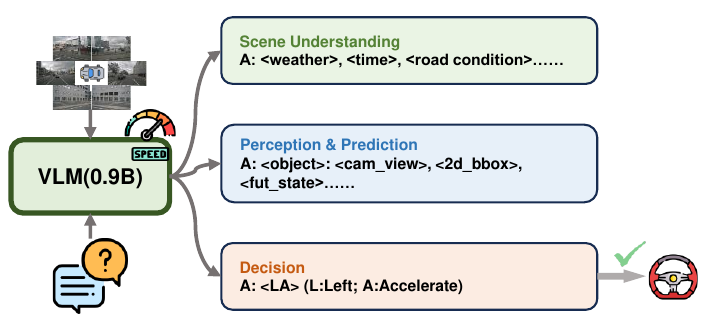}
      \caption{FastDrive (Ours)}
  \end{subfigure}
  \caption{Existing VLMs heavily rely on massive-parameter VLMs and free-form textual annotations, which introduce computational overhead and hinder inference efficiency. FastDrive, a compact VLM baseline for end-to-end autonomous driving with structured data, which enhances inference efficiency and integration into autonomous driving systems.}
  \label{fig:abstract}
\end{figure}

Current VLMs training paradigms heavily rely on datasets with free-form textual annotations (Fig.~\ref{fig:abstract}), such as NuScenes-QA \cite{qian2024} and BDD-X \cite{kim2018}. While these descriptions capture rich semantic information, their syntactic variability forces models to parse redundant linguistic patterns. For example, the sentence ``A black sedan is turning left" and ``A sedan that is black is making a left turn" convey the same information but differ in structure, this syntactic variability increases the complexity of the learning task and computational overhead, as VLMs must disambiguate synonymous expressions rather than focus on core reasoning tasks. Additionally, another sentence " Moving pedestrian wearing a white top and gray shorts in the crosswalk" along with the previous example contain redundant information, such as the color of the vehicle and the pedestrian's clothing, which could introduce unnecessary cognitive load for downstream decision-making processes in autonomous driving systems. In this context, the VLMs may spend significant attention on irrelevant information rather than focusing on core event reasoning, causing wastage of computational resources and hindering inference efficiency. Moreover, we also observe that existing some baselines often rely on large-scale VLMs, such as DriveLM \cite{drivelm}, DriveVLM \cite{drivevlm}, and LeapAD\cite{leapad}, etc, which tpically exceed 7B parameters or more to achieve multimodal alignment and reasoning. Although ultra-large parameter parameters VLMs may achieve fair performance in various benchmarks, along with the high computational cost, memory consumption, and inference latency, which rendering them impractical for real-time deployment in autonomous driving systems.

To address these challenges, this paper introduces a structured and concise benchmark dataset, NuScenes-S, derived from the NuScenes dataset \cite{caesar2020nuscenes}. Different from existing datasets that feature free-form textual annotations with redundant information, NuScenes-S extract and summarize key elements that may affect driving decisions, such as vehicle type, vehicle action, pedestrian action, traffic light status, etc., into clear and concise phrases, and organize them into structured dictionarie format. By converting key information into structured key-value pairs, it ensures data consistency and significantly reduces the computational cost associated with natural language parsing. This structured representation allows for efficient retrieval of relevant information while filtering out redundant content, thereby enhancing the clarity and relevance of the input to downstream modules. Furthermore, it allows for the flexible construction of tailored question-answer pairs, which not only facilitates targeted model training but also provides a more effective and interpretable framework for comprehensive model evaluation. Additionally, a compact VLM baseline referenced from InternVL \cite{chen2025internvl} is introduced, named FastDrive, which is designed for end-to-end autonomous driving with small-scale parameters. FastDrive mimics the reasoning strategies of human drivers by employing a chain-of-thought process to perform scene understanding, perception, prediction, and decision-making tasks, thereby achieving effective alignment with end-to-end autonomous driving frameworks. In summary, the main contributions of this paper are as follows:


\begin{itemize}
    \item We introduce a structured dataset that focuses on key elements closely related to driving decisions, which eliminates redundant information and addresses the limitation of synonymous expressions in free-form textual annotations and enhances the efficiency of inference.  
    \item A compact VLM baseline with 0.9B parameters is proposed, which mimics the reasoning strategies of human drivers and achieves effective alignment with end-to-end autonomous driving frameworks.
    \item A comprehensive evaluation and extensive experiments tailored for NuScenes-S and FastDrive are conducted. The results demonstrate the effectiveness of the proposed dataset and model, which achieves competitive performance on the NuScenes-S benchmark. 
\end{itemize}

\begin{figure*}[htbp!]
  \setlength{\abovecaptionskip}{0.1cm}
    \centering
    \includegraphics[width=1\linewidth]{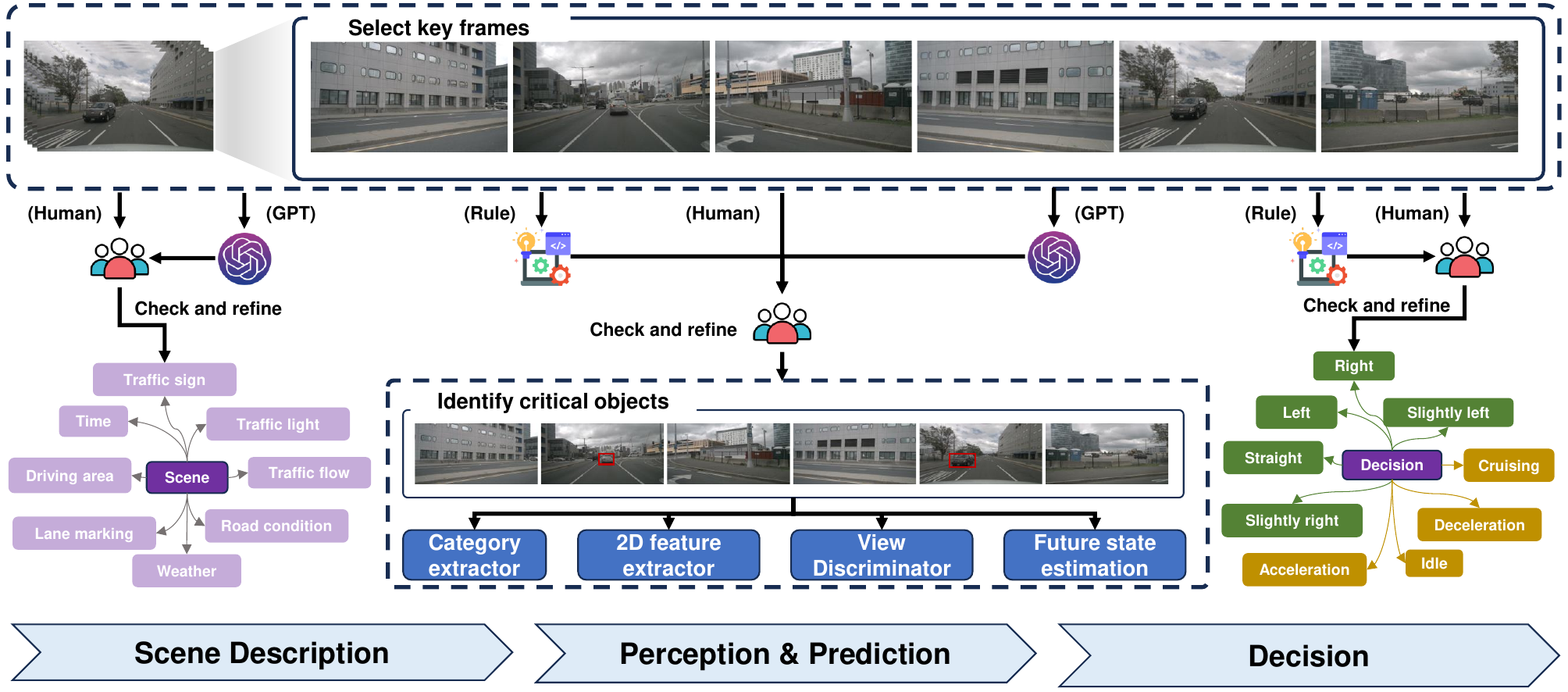}
    \caption{The dataset construction process of the NuScenes-S dataset.}
    \label{fig:data_process}
\end{figure*}


\section{Related work}
\subsection{Driving with VLMs}
VLMs \cite{liu2023visual,liu2024imp,chen2024internvl} have emerged as a transformative paradigm in artificial intelligence, bridging multimodal understanding between visual inputs and linguistic context. These models have shown remarkable performance in various vision-language tasks, such as image captioning \cite{zhu2023minigpt4}, visual question answering \cite{ding2024holi,lin2024moellava}, and visual reasoning \cite{zhang2024the}. These capabilities have enabled VLMs to understand complex visual scenes and generate plausible reasoning chains, making them a promising approach for end-to-end autonomous driving \cite{yang2024llm4drive}. Early works \cite{wen2024dilu,fu2023drivelikehumanrethinking} have explored text dialogue capabilities of LLMs for autonomous planning tasks, but relying on handcrafted rules and linguistic description makes it difficult for LLMs fully understand driving scenarios. With the advent of VLMs, more recent works use VLMs to interact directly with driving environment through visual and linguistic inputs \cite{xie2025vlmsready}. LMDriver \cite{shao2023lmdrive} and DriveMLM \cite{wang2023drivemlmaligningmultimodallarge} interact with dynamic driving environment through multimodal sensor inputs and natural language commands and directly output control commands, which treats VLMs as a black box and does not provide explicit reasoning process. DriveLM \cite{drivelm} devides the driving task into perception, prediction, and planning, and uses graph visual question answering to improve the interpretability of the reasoning process. DriveVLM \cite{drivevlm} and Senna \cite{jiang2024senna} adopts a novel approach by combining VLMs with modular end-to-end driving pipeline to compensate the shortcomings of modular black-box pipeline. LeapAD \cite{leapad} inspired by human cognitive process, introduces three different VLMs to mimic scene understanding, decision making, and reflection process. Despite the progress, existing VLMs are not impractical for real-time inference or deployment in autonomous systems due to their large number of parameters and unstructured language descriptions in existing datasets.


\subsection{Driving Datasets for VLMs}
General-purpose VLMs struggle to achieve satisfactory performance in autonomous driving tasks due to challenges in dynamic scene understanding and multimodal reasoning. Traditional autonomous driving datasets, such as KITTI \cite{doi:10.1177/0278364913491297}, Waymo Open Dataset \cite{sun2020scalabilityperceptionautonomousdriving}, and NuScenes \cite{caesar2020nuscenesmultimodaldatasetautonomous} mainly provide rich multimodal sensor data for perception or prediction tasks unsuitable for VLMs. In order to adapt to VLMs, Talk2Car \cite{Deruyttere_2019}, NuPrompt \cite{wu2023languagepromptautonomousdriving}, NuScenes-QA \cite{qian2024nuscenesqamultimodalvisualquestion} and DriveLM \cite{drivelm} introduce free-form language descriptions and QA pairs to the NuScenes dataset. BDD-X \cite{kim2018textualexplanationsselfdrivingvehicles} and BDD-OIA \cite{xu2020explainableobjectinducedactiondecision} provide text annotations describing vehicle actions and their rationales. DRAMA \cite{malla2022dramajointrisklocalization} focus on driving hazards and related objects, this dataset provides rich visual scenes and object-level queries. Rank2Tell \cite {sachdeva2023rank2tellmultimodaldrivingdataset} annotates various semantic, spatial, temporal, and relational attributes of various important objects in complex traffic scenarios. Some of these datasets mainly focus on scene understanding, risk assessment, object-level queries, or multimodal reasoning, but lack structured language descriptions and reasoning chains, which could improve VLMs' capacity for understanding driving scenarios and enhance inference efficiency. Other datasets may include fairly complete autonomous driving tasks, but the language descriptions are unstructured and verbose, which hinders the integration of VLMs into autonomous systems. Our NuScenes-S manages dataset in a human-like reasoning manner across perception, prediction, and decision-making tasks, which focus on key elements in driving scenarios and convert unstructured language descriptions into structured and concise format, further improving efficiency and integration.

\begin{table*}[htbp!]
    \setlength{\abovecaptionskip}{0.1cm}
  \centering
  \tabcolsep=0.2cm
  \renewcommand{\arraystretch}{0.8}

  \begin{tabular}{@{}l|cccccccc|ccc|cc|c@{}}
    \toprule
    \multirow{2}{*}{\textbf{Dataset}} & \multicolumn{8}{c|}{\textbf{Scene Description}} & \multirow{2}{*}{\textbf{Per.}} & \multirow{2}{*}{\textbf{Pre.}} & \multirow{2}{*}{\textbf{Dec.}} & \textbf{Frames} & \textbf{QA Pairs} & \multirow{2}{*}{\textbf{Format}}\\
    & \textbf{wea.} & \textbf{time.} & \textbf{con.} & \textbf{road.} & \textbf{area.} & \textbf{mark.} & \textbf{light.} & \textbf{sign.} &&&& (Test) & (Test)\\
    \midrule
    BDD-X & \ding{55} & \ding{55} & \ding{55} & \ding{55} & \ding{55} & \ding{55} & \ding{55} & \ding{55} & \usym{1F5F8} & \ding{55} & \ding{55} & - & - & textual \\
    BDD-OIA &\ding{55} & \ding{55} & \ding{55} & \ding{55} & \ding{55} & \ding{55} & \ding{55} & \ding{55} & \usym{1F5F8} & \ding{55} & \usym{1F5F8} & - & - & textual \\
    NuScenes-QA & \ding{55} & \ding{55} & \ding{55} & \ding{55} & \ding{55} & \ding{55} & \ding{55} & \ding{55} & \usym{1F5F8} & \ding{55} & \ding{55} & 36114 & 83337 & textual \\
    Talk2Car &\ding{55} & \ding{55} & \ding{55} & \ding{55} & \ding{55} & \ding{55} & \ding{55} & \ding{55} & \usym{1F5F8} & \ding{55} & \usym{1F5F8} & 1.8k & 2447 & textual \\
    nuPrompt & \ding{55} & \ding{55} & \ding{55} & \ding{55} & \ding{55} & \ding{55} & \ding{55} & \ding{55} & \usym{1F5F8} & \ding{55} & \ding{55} & 36k & 6k & textual \\
    DRAMA &\ding{55} & \ding{55} & \ding{55} & \ding{55} & \ding{55} & \ding{55} & \ding{55} & \ding{55} & \usym{1F5F8} & \ding{55} & \usym{1F5F8} & - & - & textual \\
    Rank2Tell & \ding{55} & \ding{55} & \ding{55} & \ding{55} & \ding{55} & \ding{55} & \ding{55} & \ding{55} & \usym{1F5F8} & \ding{55} & \usym{1F5F8} & - & - & partially structured \\
    DriveLM & \ding{55} & \ding{55} & \ding{55} & \ding{55} & \ding{55} & \ding{55} & \ding{55} & \ding{55} & \usym{1F5F8} & \usym{1F5F8} & \usym{1F5F8} & 4794 & 15480  & partially structured\\
    DriveVLM & \usym{1F5F8} & \usym{1F5F8} & \ding{55} & \usym{1F5F8} & \ding{55} & \ding{55} & \ding{55} & \ding{55} & \usym{1F5F8} & \usym{1F5F8} & \usym{1F5F8} & - & - & partially structured \\
    NuScenes-S & \usym{1F5F8} & \usym{1F5F8} & \usym{1F5F8} & \usym{1F5F8} & \usym{1F5F8} & \usym{1F5F8} & \usym{1F5F8} & \usym{1F5F8} & \usym{1F5F8} & \usym{1F5F8} & \usym{1F5F8} & 6019 & 18057 & structured \\
    \bottomrule
  \end{tabular}
  \caption{ Comparison among benchmark datasets for autonomous driving. }
  \label{tab:dataset_comparion}
\end{table*}


\section{The NuScenes-Structured Benchmark}
\label{sec:benchmark}

\subsection{Scene Description}
\label{sec:scene_description}
Understanding the driving scenario is crucial for making safe driving decisions. Therefore, the scene description in NuScenes-S is introduced to provide a more comprehensive view of the driving scenario, addressing the often overlooked or insufficiently represented aspects in many existing datasets \cite{drivelm,sachdeva2023rank2tellmultimodaldrivingdataset,Deruyttere_2019,wu2023languagepromptautonomousdriving,malla2022dramajointrisklocalization}. The scene description in NuScenes-S is structured and concise, which includes the following key elements: \{ \textit{Weather, Traffic condition, Driving area, Traffic light, Traffic sign, Road condition, Lane markings, Time}\}.


\begin{itemize}
  \item {\textbf{Weather}}: Weather conditions play a crucial role in driving, as adverse weather can reduce visibility and alter road conditions, ultimately leading to more cautious driving decisions. Weather conditions include \{\textit{sunny, rainy, snowy, foggy, cloudy}\}. 
  \item {\textbf{Traffic condition}}: Different traffic conditions will introduce different driving challenges, traffic congestion will affect the driver's speed and decision-making. Traffic conditions include \{\textit{low, moderate, moderate}\}.
  \item {\textbf{Driving area}}: Each driving area has its own characteristics, such as intersections and junctions that pose more challenges for turning and lane changing decisions. Driving areas include \{\textit{intersection, junction, roundabout, residential, crosswalk, parking lot}\}.
  \item {\textbf{Traffic light}}: Traffic lights are important traffic control devices that regulate the flow of traffic. The state of traffic lights will affect the driver's decision-making. Traffic lights  include \{\textit{green, yellow, red}\}.
  \item {\textbf{Traffic sign}}: Traffic signs provide important information for drivers changing driving behavior to follow the rules. Traffic signs include \{\textit{speed limit, stop, yield, no entry, no parking, no stopping, no u-turn, no left turn, no right turn, no overtaking, one way}\}.
  \item {\textbf{Road condition}}: Road conditions are critical for driving safety: construction zones require caution, and wet or icy roads require slower speeds and longer following distances. Road conditions include \{\textit{smooth, rough, wet, icy, construction}\}.
  \item {\textbf{Lane markings}}: Lane markings provide directional guidance to guide drivers' driving decisions. Lane markings include \{\textit{right turn, left turn, straight, straight and right turn, straight and left turn, u-turn, left and u-turn, right and u-turn}\}.
  \item {\bf Time}: Time represents the time of day, driver tends to drive more cautiously at night due to reduced visibility. Time includes \{\textit{daytime, night}\}.
\end{itemize}

\begin{figure}
    \setlength{\abovecaptionskip}{0.1cm}
  \centering
  \includegraphics[width=1\linewidth]{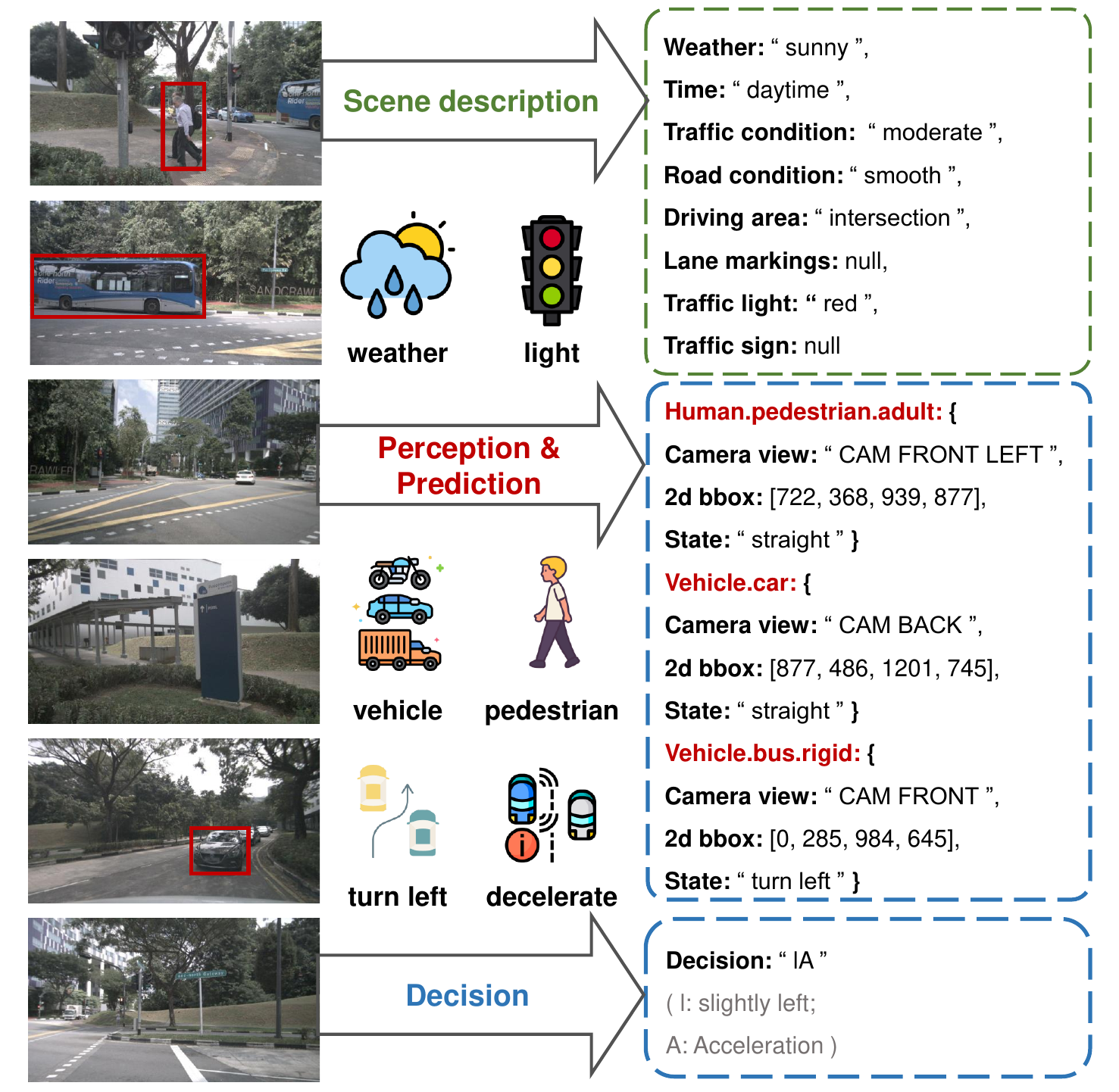}
  \caption{An annotation example of the NuScenes-S dataset.}
  \label{fig:data_example}
\end{figure}

\begin{figure*}
    \setlength{\abovecaptionskip}{0.1cm}
    \centering
    \includegraphics[width=0.9\linewidth]{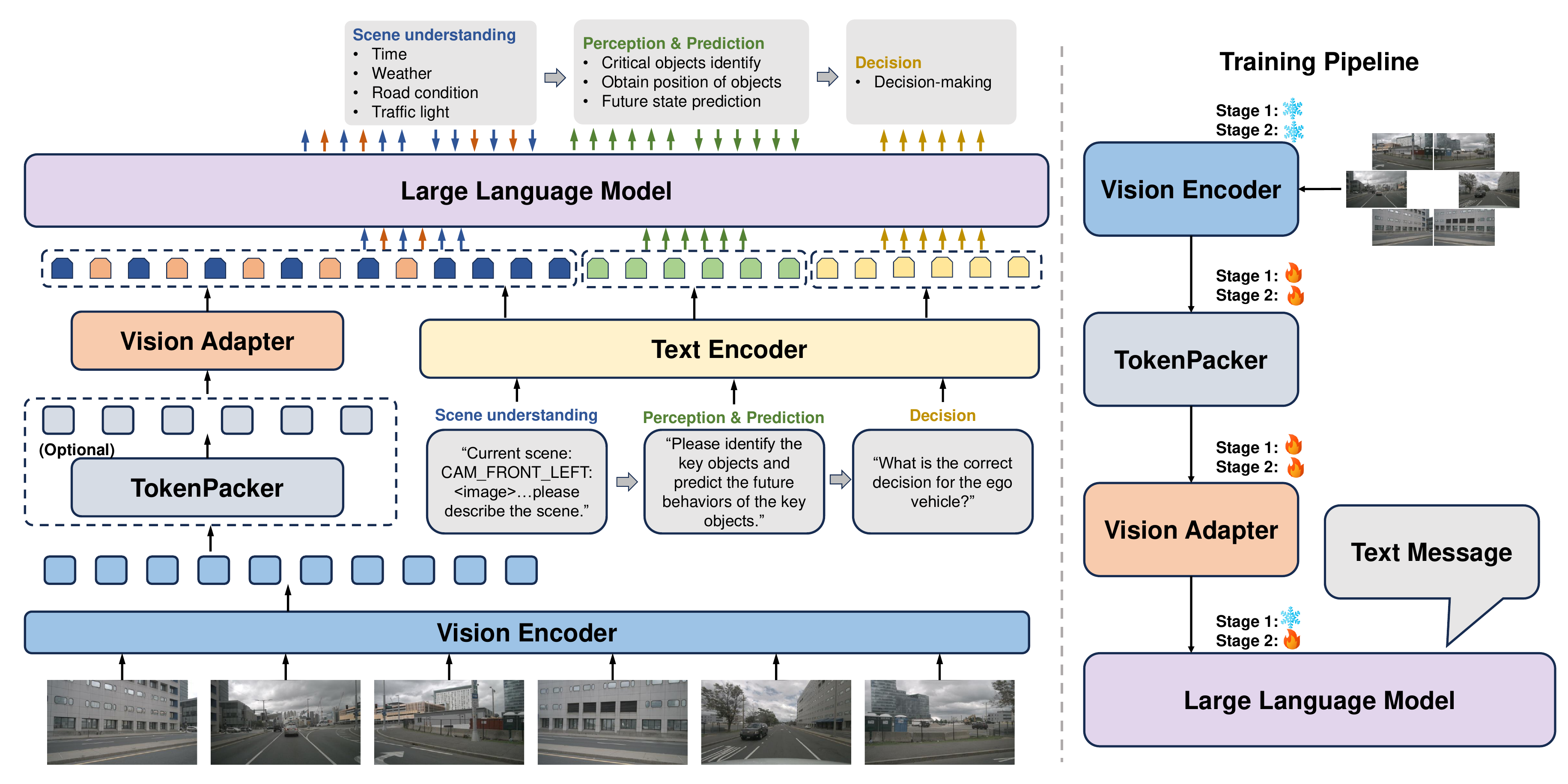}
    \caption{The framework of the FastDrive model for end-to-end autonomous driving.}
    \label{fig:framework}
\end{figure*}

\vspace{-5pt}
\subsection{Perception \& Prediction}
\label{sec:perception_prediction}
Identify some key objects and predict their future states are essential for a driver to make decisions. Most existing datasets \cite{drivevlm,sachdeva2023rank2tellmultimodaldrivingdataset} describe these tasks in free-form language descriptions, which usually use a very long and verbose sentence or paragraph to describe a perception or prediction task while truly contains only a few key elements. To address this issue, we incorporated the perception and prediction tasks into the NuScenes-S dataset and managed them in a structured and concise manner to improve the efficiency and integration of VLMs. The perception and prediction tasks in NuScenes-S are structured as follows: \{\textit{Object:}\{\textit{Camera view, 2D bounding box, Future state}\}\}. 

\begin{itemize}
  \item {\textbf{Object}}: The object is the key element in the perception and prediction tasks, which includes the following attributes: \{\textit{Camera view, 2D bounding box, Future state}\}.
  \item {\textbf{Camera view}}: The camera view of the object, which helps ego vehicle to identify the direction of the object in decision making. The camera view includes \{\textit{Front, Front left, Front right, Back, Back left, Back right}\}.
  \item {\textbf{2D bounding box}}: The 2D bounding box of the object, which helps ego vehicle to locate the object in the camera view. The 2D bounding box consists of coordinates of the two diagonal vertices \{\textit{x1, y1, x2, y2}\}.
  \item {\textbf{Future state}}: The future state of the object, the ego vehicle makes driving decisions based on the future state of the object. The future state includes \{\textit{Straight, Turn left, Turn right, Slightly left, Slightly right, Stop, Idle}\}. 
\end{itemize}


\subsection{Decision}
\label{sec:decision}
Make decisions based on the perception and prediction tasks is the final and critical step for a driver to drive safely. Current method rely on linguistic descriptions to describe the decision-making process that limit the integration of VLMs into autonomous systems. To address this issue, we treat the decision-making task as visual action reasoning thus convert the decision-making task into VLA task through defining some ruled-based actions similar to modular driving system. The decision-making task in NuScenes-S is structured as follows: \{\textit{Decision:} \{\textit{Lateral movement, Longitudinal movement}\}\}.
\begin{itemize}
  \item {\textbf{Decision}}: The decision is a safe driving action that the ego vehicle could take based on the perception and prediction tasks, which includes the following attributes: \{\textit{Lateral movement, Longitudinal movement}\}.
  \item {\textbf{Lateral movement}}: The lateral movement of the vehicle, which includes \{\textit{Turn left(L), Turn right(R), Slightly left(l), Slightly right(r), Straight(S)}\}.
  \item {\textbf{Longitudinal movement}}: The longitudinal movement of the vehicle, which includes \{\textit{Accelerate(A), Decelerate(D), Cruising(C), Idle(I)}\}.
\end{itemize}


\subsection{Dataset Construction}
\label{sec:dataset_construction}
In order to construct a high-quality structured benchmark dataset, we construct the datasets with a tiered and comparative optimization manner through holistic integration of rule-based annotation, VLM annotation, and human refinement, as is shown in Fig.~\ref{fig:data_process}. Specifically, in scene description, we first annotation scene information through GPT and human annotators, then we use compare the results of GPT and human annotators to find the difference and refine the annotations by human annotators. Similarly, in perception and prediction tasks, we first define some rules to extract key objects  then we use VLMs and human annotators to annotation the key objects synchronously. Subsequently, through comparative optimization and human refinement to ensure the quality of the dataset. Finally, the related information of key objects could be extracted directly from NuScenes dataset. Finally, The decision task is annotated rule-based and human annotators to get initial annotations, then further refined by human annotators with comparative optimization. It is worth noting that by strategically organizing the annotation sequence, partial parallelization of the annotation tasks can be achieved, thereby improving annotation efficiency. On the other hand, by combining multiple annotation methods and employing contrastive optimization, the arbitrariness of relying on a single annotation method is avoided, further enhancing the quality of the dataset.

\section{FastDrive}
\label{sec:driver-s}
The overview of the FastDrive is shown in Fig.~\ref{fig:framework}. FastDrive is a compact VLM for end-to-end autonomous driving with parameters of 0.9B, significantly lower than current methods.
The model follows the ``ViT-Adapter-LLM" architecture widely used in various MLLM studies \cite{liu2023visual,liu2024imp} but introduced an optional TokenPacker module that reduce the number of visual tokens to improve the inference speed. Moreover, we fine tune the model by chaining autonomous driving tasks into a reasoning process, aiming to accelerate the model's learning of the relationships between these tasks and improve the model's performance on the NuScenes-S benchmark.


\subsection{Vision Encoder}
The backbone of the Vision Encoder is a Vision Transformer (ViT) based on Intern ViT-300M \cite{chen2025expandingperformanceboundariesopensource}, which is distilled from the teacher model Intern ViT-6B \cite{chen2024internvl}. The ViT backbone consists of a stack of 24 Transformer blocks with 16 heads, and the hidden size of the model is 1024 with 0.3B parameters. It can achieve a competitive performance on various vision-language tasks while maintaining a relatively small number of parameters by incrementally pre-training the model on large-scale datasets. As is shown in Fig.~\ref{fig:framework}, the ViT backbone takes the input images of six different views, including front, front left, front right, back, back left, and back right, and extracts visual features from the images. The visual features are then projected into feature space of LLM by an MLP adapter. Additionally, we introduce an optional TokenPacker module to reduce the number of visual tokens, which can improve the inference speed of the model while maintaining the competitive performance.
\subsection{LLM Agent}
As illustrated in Fig.~\ref{fig:framework}, the LLM plays a "brain" role in the FastDrive model throughout the driving process, which takes the visual features from the Vision Encoder and the structured language instructions as input and generates the scene description, identifys key objects, predicts their future states, and makes driving decisions in a chain of thought (CoT) manner. Specifically, we choose Qwen2.5 \cite{qwen2025qwen25technicalreport} as the LLM agent in the FastDrive model, which is a small LLM model with 0.5B parameters. Qwen2.5 has shown competitive performance on a wide range of benchmarks evaluating language understanding, reasoning, etc. It has achieved a significant improvement in instruction following, understands structured data and generates structured outputs. Thus, we select Qwen2.5 as the LLM agent in the FastDrive baseline model.


\section{Experiments}
\label{sec:experiments}

\subsection{Implementation Details}
\label{sec:experimental_settings}
The experiments are conducted on 8 NVIDIA RTX 4090 GPUs. The FastDrive model is trained with a batch size of 1 for 10 epochs using the Adam optimizer with an initial learning rate of 1e-4, and the learning rate is decayed by a factor of 0.05. 
The experiments are conducted on the NuScenes-S dataset, which contains about 102K QA pairs in total. The dataset is split into 84K training QA pairs, 18K test QA pairs. 
The evaluation metrics include Language metrics, Average Precision (AP), Recall, Precision and Decision Accuracy for perception, prediction, and decision-making tasks.

\begin{table*}[htbp!]
  \centering
  \tabcolsep=0.17cm
  \renewcommand{\arraystretch}{1}
  \begin{tabular}{@{}l|cccccc|cccccccc@{}}
    \toprule
    \multirow{2}{*}{\textbf{Method}} & \multicolumn{6}{c|}{\textbf{Language}} & \multicolumn{8}{c}{\textbf{Accuracy} (\%)}\\
    \cmidrule{2-7}
    \cmidrule{8-15}
    & \textbf{BLEU\_1} & \textbf{BLEU\_2} & \textbf{BLEU\_3} & \textbf{BLEU\_4} & \textbf{ROUGE\_L} & \textbf{CIDEr} & \textbf{weather} & \textbf{time} & \textbf{traffic} & \textbf{road} & \textbf{area} & \textbf{mark} & \textbf{light} & \textbf{sign}\\
    \midrule
    DriveLM& 82.70 & 76.51 & 70.41 & 65.05 & 83.93 & 5.30 & 85.47 & 99.91 & 76.30 & 83.85 & 74.96 & 81.49 & 85.57 & 83.90 \\
    FastDrive\textsubscript{64}  & 80.49 & 77.66 & 72.77 & 68.06 & 60.53 & 3.58 & 93.35 & 99.81 & 78.08 & 86.57 & 75.98 & \textbf{82.31} & \textbf{88.22} & 85.85 \\
    FastDrive\textsubscript{256} & \textbf{86.77} & \textbf{81.09} & \textbf{75.34} & \textbf{70.36} & \textbf{87.24} & \textbf{6.20} & \textbf{94.13} & \textbf{99.95} & \textbf{78.15} & \textbf{87.66} & \textbf{76.49} & 82.06 & 87.74 & \textbf{87.64} \\
    \bottomrule
  \end{tabular}
  \caption{ Performance of scene description on the NuScenes-S dataset. \textbf{Bold} indicates the best performance. FastDrive\textsubscript{64} (with TokenPacker) and FastDrive\textsubscript{256} are the FastDrive models with 64 and 256 tokens, respectively. The same applies to the following tables.}
  \label{tab:scene_description}
\end{table*}

\begin{table*}[htbp!]
        \setlength{\abovecaptionskip}{0.1cm}
  \centering
  \tabcolsep=0.21cm
  \begin{tabular}{@{}lcccccccc|c|cccc@{}}
    \toprule
    \multirow{2}{*}{\textbf{Method}} & \multicolumn{8}{c|}{\textbf{Perception}} & \textbf{Prediction}& \multicolumn{4}{c}{\textbf{Decision}}\\
    \cmidrule{2-14}
    & \textbf{BLEU\_1} & \textbf{BLEU\_2} & \textbf{BLEU\_3} & \textbf{BLEU\_4} & \textbf{ROUGE\_L} & \textbf{CIDEr} & \textbf{AP} & \textbf{Recall} & \textbf{State} & \textbf{Dec} & \textbf{Dec(s)}&\textbf{Lat} & \textbf{Lon}\\
    \midrule
    DriveLM & \textbf{34.82} & \textbf{29.59} & \textbf{23.23} & \textbf{17.45} & \textbf{35.31} & 0.74 & 0.21 & 0.30 & 0.36 & 0.28 & 0.59 & 0.72 & 0.35\\
    FastDrive\textsubscript{64} & 26.07 & 15.17 & 8.86 & 4.25 & 34.37 &\textbf{0.75} & 0.31 & 0.45 & \textbf{0.44} & 0.38 & \textbf{0.63} & 0.74 & 0.45\\
    FastDrive\textsubscript{256} & 26.48 & 15.23 & 9.11 & 4.75 & 34.77 &0.61 & \textbf{0.37} & \textbf{0.53} & \textbf{0.44} & \textbf{0.39} & \textbf{0.63} & \textbf{0.76} & \textbf{0.46}\\
    \bottomrule
  \end{tabular}
  \caption{Performance of perception, prediction, and decision-making tasks on the NuScenes-S dataset.  DEC represents the accuracy of decision results that are consistent with the ground truth. Dec(s) represents the proportion of safe decisions, including those that match the ground truth as well as those that deviate from the ground truth but are still considered safe.}
  \label{tab:perception_prediction}
\end{table*}

\begin{table}[htbp!]
  \setlength{\abovecaptionskip}{0.1cm}
  \centering
  \setlength{\tabcolsep}{0.35cm} 
  \begin{tabular}{@{}lcccc@{}}
    \toprule
    \textbf{Method} & \textbf{Params} & \textbf{Trainable} & \textbf{Memory (GB)} & \textbf{FPS} \\
    \midrule
    DriveLM\textsuperscript{$\ddagger$} & 3.955B & 12.9M & 14.43 & 0.20 \\
    DriveLM & 3.955B & 12.9M & 14.43 & 0.36 \\
    FastDrive\textsubscript{64}\textsuperscript{$\ddagger$} & 0.9B & 8.79M & 1.97 & 2.86 \\
    FastDrive\textsubscript{256}\textsuperscript{$\ddagger$}& 0.9B & 8.79M & 1.97 & 2.11 \\
    FastDrive\textsubscript{64} & 0.9B & 8.79M & 1.97 & 4.85 \\
    FastDrive\textsubscript{256} & 0.9B & 8.79M & 1.97 & 4.01 \\
    \bottomrule
  \end{tabular}
  \caption{Comparison of model parameters, trainable parameters, FLOPs, and inference speed (FPS) for different models. $\ddagger$ indicates that the model is tested on the DriveLM dataset.}
  \label{tab:inference_speed}
\end{table}

\begin{table}[htbp!]
        \setlength{\abovecaptionskip}{0.1cm}
  \centering
  \setlength{\tabcolsep}{0.21cm} 
  \begin{tabular}{@{}lcccc|cccc@{}}
    \toprule
    \multirow{2}{*}{\textbf{Scene}} & \multicolumn{4}{c|}{\textbf{FastDrive}} & \multicolumn{4}{c}{\textbf{FastDrive w/o}} \\
    \cmidrule{2-9}
                    & \textbf{Dec}  & \textbf{Dec(s)} & \textbf{Lat} & \textbf{Lon} & \textbf{Dec}  & \textbf{Dec(s)} & \textbf{Lat} & \textbf{Lon}\\
    \midrule
    weather & 0.35 & 0.55& 0.79 & 0.44 & 0.33 & 0.51 & 0.78 & 0.43 \\
    time & 0.38 & 0.59 & 0.84 & 0.45 & 0.37 & 0.58 & 0.86 & 0.43 \\
    traffic & 0.44 & 0.69 & 0.85 & 0.47 & 0.40 & 0.66 & 0.82 & 0.47 \\
    road & 0.40 & 0.64 & 0.78 & 0.47 & 0.39 & 0.64 & 0.78 & 0.46 \\
    area & 0.33 & 0.53& 0.66 & 0.44& 0.28 & 0.51 & 0.66 & 0.39 \\
    mark & 0.39 & 0.64 & 0.77 & 0.47 & 0.39 & 0.63 & 0.77 & 0.46 \\
    light & 0.39 & 0.69 & 0.57 & 0.48 & 0.42 & 0.63 & 0.68 & 0.51 \\
    sign & 0.42 & 0.65 & 0.68 & 0.53 & 0.40 & 0.64 & 0.70 & 0.51 \\
    \bottomrule
  \end{tabular}
  \caption{Ablation studies on the impact of scene annotations on driving decisions. w/o indicates the ablation study.}
  \label{tab:ablation_studies}
\end{table}

\begin{figure*}
    \setlength{\abovecaptionskip}{0.1cm}
    \centering
    \includegraphics[width=1\linewidth]{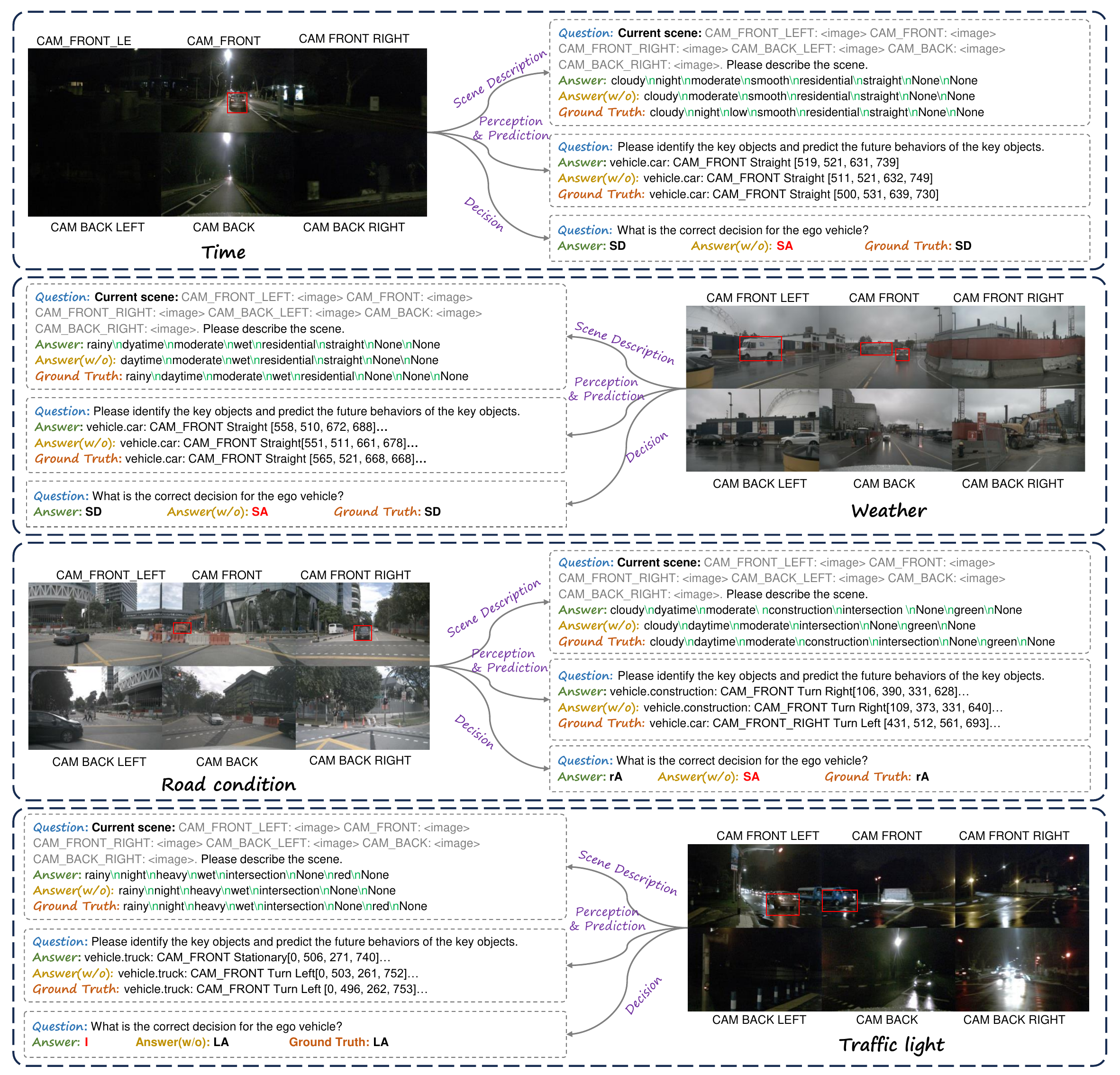}
    \caption{Examples of ablation studies on the impact of scene annotations on driving decisions. The red decision represents a decision that is not consistent with the ground truth.}
    \label{fig:ablation_example}
\end{figure*}

\subsection{Quantitative Results}
\label{sec:experimental_results}

\noindent
\textbf{Scene Understanding.} The TABLE.~\ref{tab:scene_description} show the performance of scene understanding on the NuScenes-S dataset, the results demonstrate that the FastDrive model achieves competitive performance on the structured benchmark dataset. In TABLE.~\ref{tab:perception_prediction}, we compare the models' performance in perception, prediction, and decision-making tasks. The DriveLM model excels in perception with higher language metrics but lower accuracy, while FastDrive outperforms in prediction and decision-making with higher accuracy. This raises the issue that language evaluation metrics may not be suitable for assessing autonomous driving tasks, as they primarily measure fluency and coherence, which are important for natural language processing but do not capture the practical aspects of autonomous driving. These metrics focus on how well the generated text flows or aligns with human expectations, but they fail to evaluate the model's functional correctness in decision-making, perception, and real-world performance. In autonomous driving, what matters most is how effectively the model interprets sensory data and makes safe, accurate driving decisions, aspects that may be challenging to fully capture with verbose language descriptions.

\noindent
\textbf{Perception \& Prediction \& Decision.} Additionally, it's worth noting that the language evaluation metrics get worse in perception tasks, which may raise another current VLMs may further improve the reasoning capabilities since the perception tasks are more challenging with complex and multimodal reasoning compared to the scene understanding. Moreover, the final task of end-to-end autonomous driving is generating safe and reasonable driving decisions, which is the most critical and challenging task for VLMs. From the TABLE.~\ref{tab:perception_prediction}, the FastDrive model achieves the best performance in decision-making tasks with the highest accuracy metrics. However, we also observed that the decision accuracy is relatively low. Further analysis revealed that the proportion of safe decisions is relatively high, indicating that the Vision-Language Model (VLM) tends to favor more conservative decisions. In addition, the accuracy of lateral (horizontal) decisions is higher than that of longitudinal (vertical) decisions, reflecting that longitudinal decision-making may be inherently more challenging.

\noindent
\textbf{Inference Acceleration} We conduct comparative latency analysis across models in TABLE.~\ref{tab:inference_speed}. Experimental results demonstrate that FastDrive achieves 4.85 FPS inference speed while maintaining competitive performance on the NuScenes-S benchmark, representing a 13.5$\times$ acceleration over DriveLM's 0.36 FPS baseline. This efficiency stems from three synergistic optimizations: (1) Architectural compactness reduces computational overhead (0.9B vs. 3.96B parameters); (2) Systematic conversion of unstructured linguistic inputs into structured formats via NuScenes-S, eliminating redundant semantic processing; (3) Visual token compression reduces the number of visual tokens, further improving inference efficiency. While current implementation employs basic token pruning strategies, advanced visual compression architectures present promising directions for future investigation.


\subsection{Ablation Studies}
\label{sec:ablation_studies}

To evaluate the impact of scene annotation information on driving decisions, we design a comprehensive set of ablation experiments to observe how the absence of each factor influences the model's decision-making performance. Specifically, we perform a series of fine-tuning experiments, where we systematically remove individual types of scene annotation elements. Then we compare the performance of these ablated models with the fully annotated model in corresponding challenging scenarios, providing a detailed analysis of how different types of scene information contribute to the model's decision-making capabilities. The results are shown in TABLE.~\ref{tab:ablation_studies} and Fig.~\ref{fig:ablation_example}. The results show that the FastDrive model with complete scene annotations achieves better performance in driving decisions than the FastDrive model without specific scene annotations in challenging scenarios, which indicates that the scene annotations are beneficial for the model to make safe and reasonable driving decisions. 

It's worth noting that in the traffic light ablation experiment, the ablated model slightly outperformed the complete model. This can be attributed to the logical complexity introduced by traffic lights and the conservative nature of the model. From the results, we observe that the model more tends to adopt overly conservative decisions when traffic light information is provided. As is illustrated in Fig.~\ref{fig:ablation_example}, the ego vehicle tends to turn left when the traffic light is red, which is a safe and reasonable drive decision. The model tends to adopt overly conservative decisions to ensure safety when capture the traffic light information while the ablation model relies more directly on dynamic scene context and the behavior of surrounding traffic participants, allowing it to make decisions that align more closely with the actual ground truth. In all, the results show that traffic lights do indeed impact the model's decision-making, highlighting a potential research direction for the efficient integration of scene information to strike a balance between safety and accuracy in autonomous driving.


\section{Conclusion}  
\label{sec:conclusion}
\sloppy
In this work, we introduce the NuScenes-S dataset, a structured benchmark dataset for autonomous driving, which follows the human-like reasoning process across perception, prediction, and decision-making tasks. The NuScenes-S address the limitations of redundancy and synonymous expressions caused by free-form and lengthy language descriptions in existing datasets through structured labeling. This approach reduces the complexity of handling unstructured information, allowing the model to process and interpret data more effectively, leading to more efficient decision-making. We also present the FastDrive, a compact VLM for end-to-end autonomous driving, which achieves competitive performance on the NuScenes-S dataset with faster inference speed and fewer parameters on NuScenes-S dataset. This highlights the potential of small-parameter models in structured benchmark datasets. Moreover, we conduct extensive experiments analysis the impact of scene annotations on driving decisions, which demonstrates that the scene annotations are beneficial for the model to make safe and reasonable driving decisions. We believe that the NuScenes-S dataset and the FastDrive model will serve as a valuable resource for future research in autonomous driving and structured benchmark datasets.


\bibliographystyle{IEEEtran} 
\bibliography{IEEEabrv,IEEEexample}

\end{document}